%% file: IcoSoku.tex
\RequirePackage{amsmath}
\documentclass[envcountsame]{llncs}
\usepackage[title]{appendix}
\usepackage{llncsdoc}
\usepackage{silence}
\usepackage[utf8]{inputenc}
\usepackage[T1]{fontenc}
\usepackage{amsfonts}
\usepackage{graphicx}
\usepackage{xspace}
\usepackage{tabularx}
\usepackage{fancyhdr}
\usepackage{listings}
\usepackage{changepage}
\usepackage[utf8]{inputenc}
\usepackage{filecontents}
\usepackage{pifont}
\usepackage{romannum}
\usepackage{graphicx}
\usepackage[table]{xcolor}
\usepackage{flexisym}
\usepackage{adjustbox}
\usepackage{tikz}
\usepackage{diagbox}
\usepackage{multirow}
\usepackage{tikzscale}
\usepackage{filecontents}
\usepackage{textcomp}
\usepackage{eqnarray}
\usepackage{nicefrac}
\usepackage{silence}
\usepackage{hhline}
\usepackage{siunitx}
\usepackage{hyperref}
\usepackage{enumitem,kantlipsum}
\usepackage{color, colortbl}
\usepackage{hhline}
\usepackage{float}
\definecolor{MyGray}{gray}{0.85}
\usetikzlibrary{positioning,arrows,decorations.markings,automata}
\newcolumntype{C}[1]{>{\centering\arraybackslash}p{#1}}

\begin{document}
\pagenumbering{arabic}
\title{Exploring Properties of Icosoku by Constraint Satisfaction Approach}
\titlerunning{}
\author{Ke Liu \href{https://orcid.org/0000-0002-5256-9253}{\includegraphics[scale=0.5]{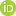}}, Sven L{\"o}ffler, and Petra Hofstedt}
\institute{Brandenburg University of Technology Cottbus-Senftenberg, Germany
\\Department of Mathematics and Computer Science, MINT
\\Konrad-Wachsmann-Allee 5, 03044 Cottbus \\ 
\email{\{liuke,sven.loeffler,hofstedt\}@b-tu.de}}
\maketitle
\begin{abstract}
\input{src/Abstract}
\end{abstract}
\input{src/Intro.tex}

\input{src/Background.tex}
\input{src/Model.tex}
\input{src/Experiment.tex}
\input{src/Conclusion.tex}
\medskip
\bibliographystyle{splncs04}
\bibliography{IcoSoku}
\end{document}

%% file: src/Abstract.tex
Icosoku is a challenging and interesting puzzle that exhibits highly symmetrical and combinatorial nature.
In this paper, we pose the questions derived from the puzzle, but with more difficulty and generality. In addition, we also present a constraint programming model for the proposed questions, which can provide the answers to our first two questions. The purpose of this paper is to share our preliminary result and problems to encourage researchers in both group theory and constraint communities to consider this topic further.
\keywords{Constraint programming  \and  Group theory \and Constraint modelling \and Icosoku.}

%% file: src/Intro.tex
\section{Introduction}\label{intro}
\begin{figure}[h]
  \centering
  \includegraphics[scale=0.12]{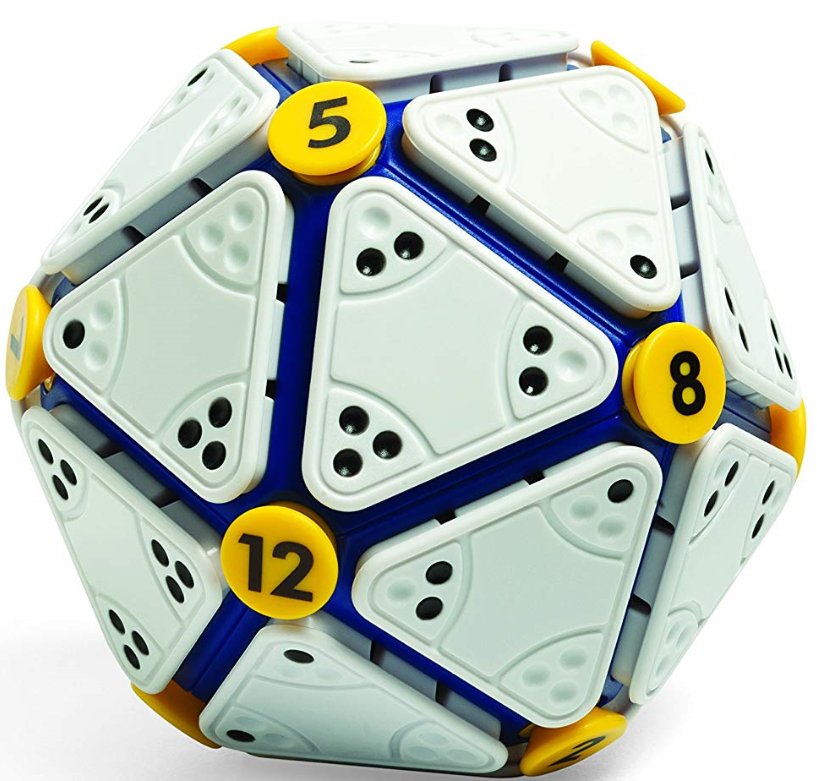}
  \caption{An icosoku~(Figure reproduced from \href{https://www.amazon.com/Recent-Toys-IcoSoKu-Brainteaser-Puzzle-x/dp/B003AIKTPU}{Amazon.com}.)}
  \label{fig_icosoku}
\end{figure}

Icosoku is a three-dimensional puzzle on a regular icosahedron block consisting of 20 tiles and 12 pegs (see Fig.~\ref{fig_icosoku}), where every vertex of a triangular tile has four possible number of black dots $\left\{0, \ldots, 3 \right\}$ and each peg takes on distinct values from  $\left\{1, \ldots, 12 \right\}$. To solve the puzzle, one needs to arrange the pegs and place the tiles. A feasible solution of the puzzle is that the value of any peg on the icosahedron is equal to the number of black dots surrounding itself. For example, the numeral 12 is surrounded by 12 black dots in Figure~\ref{fig_icosoku}.
\par
\begin{figure}[h]
\begin{center}
\begin{tikzpicture}[scale=.9, transform shape]
\tikzstyle{every node} = []
\node (a) at (0.25, 0.18) {1};
\node (b) at (1.73, 0.18) {3};
\node (c) at (1,1.44) {2};
\draw (0,0) node[anchor=north]{}
  -- (2,0) node[anchor=north]{}
  -- (1,1.73205) node[anchor=south]{}
  -- cycle;
\node (d) at (3.25, 0.18) {3};
\node (e) at (4.73, 0.18) {2};
\node (f) at (4,1.44) {1};  
\draw(3,  0) node[anchor=north]{}
  -- (5,  0) node[anchor=north]{}
  -- (4,  1.73205) node[anchor=south]{}
  -- cycle;
\node (g) at (6.25, 0.18) {2};
\node (h) at (7.73, 0.18) {1};
\node (i) at (7,1.44) {3};
\draw(6,  0) node[anchor=north]{}
  -- (8,  0) node[anchor=north]{}
  -- (7,  1.73205) node[anchor=south]{}
  -- cycle;

\draw[->] (2,  0.866025) -- (3,  0.866025) node[pos=.5,sloped,above] {$r120$};
\draw[->] (5,  0.866025) -- (6,  0.866025) node[pos=.5,sloped,above] {$r120$};
\end{tikzpicture}
\end{center}
\caption{The three symmetries of a triangular tile with values \{1,2,3\}. Here, we use numbers to replace the black dots.}
\label{fig_M1}
\end{figure}
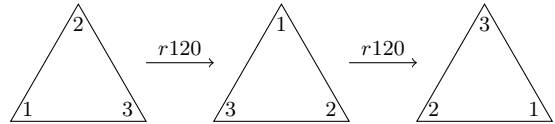

There exists $4^3$ possible triangular tiles since each of the three vertices of a triangle has four choices. However, because of the rotational symmetry, three assignments for the vertices of a triangle might represent the same triangular tile. For instance, we can rotate a tile about the triangle center by 120 and 240 degrees in a clockwise direction, as shown in Figure~\ref{fig_M1}. Therefore, there are only 24 different types of triangular tiles after breaking these symmetries.
The original icosoku puzzle only uses 14 different types of triangular tiles and claims that any arrangement of 12 pegs on the 12 vertices of the icosahedron can lead to a feasible solution. But the questions raised by the icosoku are far more than solving the puzzle itself. We believe that the icosoku is a proper research object for both constraint programming and group theory because of its combinatorial and symmetrical nature. And, thus, the following questions deserve to be explored:
\begin{enumerate}
\item Does there exist a feasible solution that the triangular tiles placed on the faces of the icosahedron are pairwise distinct? That is to say, the value at each vertex of the icosahedron is equal to the sum of values of the five vertices of the five faces that meet at this vertex of the icosahedron. And moreover, the 20 faces of the icosahedron are all different because of the values assigned to the vertices of triangular faces. 
In this paper, we call this kind of feasible solution~\textit{all different triangular solution} (ADTS).
\item Can any permutation of $\left\{ 1,\ldots, 12\right\}$ assigned to the 12 vertices of the icosahedron lead to at least one ADTS?
\item Can any 20-combination from the 24 different types of triangular tiles lead to an ADTS?
\item 
Is it possible to find a 20-combination from the 24 triangular tiles that can produce a set of feasible solutions (ADTSs) which contains all the permutations of the set $\left\{ 1,\ldots ,12\right\}$ arranged on the 12 vertices of the icosahedron?
\item 
If the answer to the previous question is affirmative, how many such 20-combinations are there?
\item How many nonisomorphic ADTSs are there if the ADTS exists?
\end{enumerate}
\par
In this paper, we present a constraint model that can answer the first two questions 
and discuss the difficulty encountered when solving the other problems. 
The rest of the paper is organized as follows. In Section~\ref{preli}, we give a brief introduction to the constraint programming. Afterward, in Section~\ref{model}, we describe our constraint model.~Then,~we present the experimental results in Section~\ref{experiment}.~Finally, we conclude in Section~\ref{conclusion}.

\par
\par

%% file: src/Background.tex
\section{Preliminaries}\label{preli}
In this section, we give some basic definitions and concepts of constraint programming (CP) and the constraints relevant to the model of the icosoku puzzle. 

The CP is a powerful technique to tackle combinatorial problems, generally NP-complete or NP-hard. A constraint satisfaction problem (CSP) can be expressed as a triple $\langle X,D,C \rangle$, where $X=\{x_1,\ldots,x_n\}$ is a set of decision variables, $D=\{D_{(x_1)},\ldots,D_{(x_n)}\}$ contains associated finite domains for each variable in $X$, and $C=\{c_1,\ldots,c_t\}$ is a collection of constraints. Each constraint $c_i$ is a relation defined over a subset of $X$, and restricts the values that can be simultaneously assigned to these variables. A solution of a CSP~\textit{P} is a complete instantiation satisfying all constraints of the CSP~\textit{P}.

The~\textbf{allDifferent} constraint is the most influential global constraint in constraint programming and widely implemented in almost every constraint solver, such as Choco solver~\cite{choco}, Gecode~\cite{schulte2010modeling}, and JaCoP~\cite{jacop}. Formally, let $X_a$ denote a subset of variables of $X$,
 the \textbf{allDifferent} constraint, which acts on $X_a$, can be defined as:
\begin{equation*}
\forall {x_i \in X_a} \forall {x_j \in X_a}(x_i \neq x_j)
\end{equation*}
The~\textbf{table} constraint is another one of the most frequently-used constraints in practice. For an ordered subset of variables $X_o=\{x_i,\ldots,x_j\} \subseteq X $, a positive (negative) \textbf{table} constraint defines that any solution of the CSP \textit{P} must (not) be explicitly assigned to a
tuple in the tuples that consists of the allowed (disallowed) combinations of values for $X_o$. For a given list of tuples $T$, we can state the positive~\textbf{table} constraint as:
\begin{equation*}
\left\{ (x_i,\ldots,x_j)  ~ \vert ~  x_i \in D_{(x_i)} ,\ldots, x_j \in D_{(x_j)} \right\} \subseteq T
\end{equation*}
The~\textbf{scalar} constraint\footnote{
This paper follows the naming convention of Choco solver. The other solvers might use a different name for the same constraint.~For instance, the \textbf{scalar} constraint is called the~\textbf{linear} and~\textbf{LinearInt} constraint in Geode and JaCoP, respectively.} is also a common global constraint, which is defined as follows:
\begin{equation*}
c_1*x_i+c_2*x_j +\ldots+c_n*x_k~\Re~sum
\end{equation*}
where $(c_1,c_2,\ldots,c_n)$ is a collection of integer coefficients, $(x_i,x_j,\ldots,x_k)$ and $sum$ are the variables on which the constraint restricts the relationship. The $\Re$ is an operator in~$ \left\{=,<,>,\neq,\leq,\geq \right\}$. Besides, the~\textbf{arithm} constraint is used to enforce relations between integer variables or between integer variables and integer values. For example, an integer value can be assigned to an integer variable by using the~\textbf{arithm} constraint. We refer to \cite{DBLP:reference/fai/2,DBLP:books/daglib/0016622,lecoutre2013constraint} for more comprehensive and profound introduction to the CP.

%% file: src/Model.tex
\section{The Constraint Programming Model}\label{model}
To solve the problem, we first should identify the decision variables for the CSP model. Then we impose constraints on these variables based on the problem definition. Focusing on the Icosoku since it has 12 vertices, a list of 12 integer variables $V=(v_1,v_2,\ldots,v_{12})$ is used to represent these vertices, each of which has domain 1..12. Since the set of values $\left\{ 1, \ldots ,12 \right\}$ has to be assigned to the 12 vertices in an ADTS, the 12 integer variables must all take distinct values. Therefore, $(v_1,v_2,\ldots,v_{12})$ must satisfy the~\textbf{allDifferent} constraint, given by:
\begin{equation}
\textrm{\textit{allDifferent}}(v_1,v_2,\ldots,v_{12})
\label{constraint4}
\end{equation}

Similarly, since there are 20 faces on a regular icosahedron, we can define a $20{\times}4$ matrix \textit{F} with integer variables for the 20 faces, where the first three elements of each row represent the three vertices of a triangular face; and the last element of each row stands for the corresponding type of the triangular tile determined by values of the first three elements of that row. For this reason, the domains of the first three columns and the last column of the matrix \textit{F} are 0..3 and 1..24, respectively. The first question posed in the Introduction (Sec.~\ref{intro}) asks whether or not a feasible solution with 20 different types of triangular tiles exist. Hence, we can also introduce the~\textbf{allDifferent} constraint to restrict that the values taken by the last column of the matrix \textit{F} are pairwise different, which can be expressed by:
\begin{equation}
\textrm{\textit{allDifferent}}(F[0,3],F[1,3],\ldots,F[19,3])
\label{constraint5}
\end{equation}
\par
As mentioned before, only 24 distinct types of triangular tiles exist after eliminating the symmetries.~However, all combinations of values that can be assigned to every row of the matrix \textit{F} are 64 \textit{4}-tuples, each of which consists of the first three values for a triangular face and the last value which indicates the type of that face. For example, as we have shown in Figure~\ref{fig_M1}, assigning the following values [(1,2,3),(3,1,2),(2,3,1)] to the three vertexes of a triangle in turn results in the same triangular tile. Thus, the tuples [(1,2,3,23),(3,1,2,23),(2,3,1,23)] contain the same type value (Table~\ref{table122}).~Because every Platonic solid has a different number of faces, we do not present the algorithm that generates all 64 tuples.
\begin{table}[h!]
\begin{center}
\begin{tabular}{p{10cm}}
\begin{equation*}
\begin{vmatrix}
0&0&0&1\\
1&1&1&2\\
2&2&2&3\\
\end{vmatrix}
\quad
\ldots
\quad
\begin{vmatrix}
0&0&2&7\\
0&2&0&7\\
2&0&0&7\\
\end{vmatrix}
\quad
\ldots
\quad
\begin{vmatrix}
0&3&3&10\\
3&0&3&10\\
3&3&0&10\\
\end{vmatrix}
\quad
\ldots
\quad
\begin{vmatrix}
1&2&3&23\\
3&1&2&23\\
2&3&1&23\\
\end{vmatrix}
\quad
\begin{vmatrix}
3&2&1&24\\
2&1&3&24\\
1&3&2&24\\
\end{vmatrix}
\end{equation*}
\end{tabular}
\end{center}
\caption{A partial list of tuples. We do not list all 64 tuples due to the limited space.}
\label{table122}
\end{table}
Let~\textit{T\textsubscript{faces}} denote the 64 tuples. We utilize the \textbf{table} constraint specified with \textit{T\textsubscript{faces}} to limit possible
combinations of values for each row of the matrix \textit{F}, which can be stated as:
\begin{equation}
I= \left\{ i \in  \mathbb{Z} \vert ~0\leq i\leq19  \right\},~\forall i \in I(\textrm{\textit{table}}(F[i,*],T_{faces}))
\label{constraint68}
\end{equation}
where $F[i,*]$ stands for a row in the matrix \textit{F}. By using these 20 table constraints, we can associate the values at the vertices of a triangular face with its corresponding type so that the Constraint~(\ref{constraint5}) can restrict the number of triangular types to be exactly 20.

The last property that an ADTS must satisfy is that the value assigned to any vertex of the icosahedron must be equal to the sum of values assigned to the vertices of the triangle surrounding this vertex of the icosahedron. To ensure this property, we can impose the \textbf{scalar} constraints on the CP model, given by:
\begin{equation}
I= \left\{ i \in  \mathbb{Z} \vert ~0\leq i\leq11  \right\},~
\forall i \in I(\textrm{\textit{scalar}}(F_{subset},\textrm{\textit{coefficients}},=,v_i))
\label{constraint6}
\end{equation}
where $F_{subset}$ is a subset of the matrix \textit{F} with cardinality five, \textit{coefficients} is an array with 5 ones, and $v_i$ denotes the decision variable for the vertices of the icosahedron. Obviously, the Constraint~(\ref{constraint6}) guarantees that $\sum F_{subset} = v_i$ where $F_{subset}$ consists of the five vertices of the five triangular faces meeting at $v_i$. Please note that we do not explicitly specify the five elements in the $F_{subset}$ because they depend on how the triangular faces and the variables representing their vertices on the icosahedron are labelled in practice.

To partially break \textit{value symmetry}~\cite{DBLP:journals/constraints/Puget05}, which preserves the solution with regard to the permutation of values, we can set the first vertex in $V$ to one. Thus, we have the constraint:
\begin{equation}
\textrm{\textit{arithm}}(v_0,=,1)
\label{constraint7}
\end{equation}
\par
In summary, Constraints~(\ref{constraint4}),~(\ref{constraint5}),~(\ref{constraint68}),~(\ref{constraint6}), and~(\ref{constraint7}) form the model used to answer the question 1 in the Introduction (Sec.~\ref{intro}). It is easy to calculate that the total number of constraints and variables are 35 and 92, respectively.

%% file: src/Experiment.tex
\section{Experiments}\label{experiment}
In this section, we present the experimental results that can answer the first two questions posed in the Introduction. We implemented the model in the Java library Choco 4.10.0~\cite{choco} running on JVM 11.0.2. All the experiments were executed on a Linux laptop with Intel i7-3720QM 2.60GHz CPU and 8 GB DDR3 memory.
\begin{table}[h]
\begin{center}
\begin{tabular}{p{2.2cm}p{1.8cm}p{2.2cm}}
Visited Nodes&Backtracks&CPU time (ms)\\
\hline
\multicolumn{1}{c}{48}&\multicolumn{1}{c}{1}&\multicolumn{1}{c}{32}
\end{tabular}
\end{center}
\caption{Result for obtaining the first ADTS}
\label{tableComparisonSmallInstance}
\end{table}
The results of our first experiment for obtaining the first ADTS is summarized in Table~\ref{tableComparisonSmallInstance}. Besides, we specified the filtering algorithms~\textit{FC} and~\textit {GAC3rm+} for all the~\textbf{allDifferent} and~\textbf{table} constraints; and the search strategy was set to the~\textit{minDomLBSearch}. 

\begin{figure}[htbp]
\begin{center}
\begin{tikzpicture}[scale=1.5]
\path (0:0cm) node (v0) {$v_0$};
\path (0:1cm) coordinate (p0);
\path (72:1cm) coordinate (p1);
\path (2*72:1cm) coordinate (p2);
\path (3*72:1cm) coordinate (p3);
\path (4*72:1cm) coordinate (p4);
\draw[thin] 
      (v0) -- (p0)
      (v0) -- (p1)
      (v0) -- (p2)
      (v0) -- (p3)
      (v0) -- (p4);
\draw[thick] (p0) -- (p1) -- (p2) -- (p3) -- (p4) -- cycle;
\draw[->,>=stealth',semithick] (150:1.2cm) arc (150:210:1.2cm);
\draw[->,>=stealth',semithick] (-30:1.2cm) arc (-30:30:1.2cm);  
\end{tikzpicture}
\end{center}
\caption{The view from the top of the vertex $v_0$ of the icosahedron}
\label{fig_M2}
\end{figure}
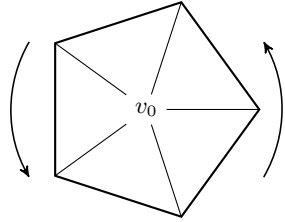

In order to answer the second question in the Introduction (Sec.~\ref{intro}), we conducted the experiment that exhaustively tests the possible permutations of the set $\{1, \dots, 12\}$ for the 12 vertices of the icosahedron by fixing the values of the $V$ in each iteration. Moreover, to reduce the computational effort, we avoid evaluating the symmetries that are generated by rotating about the vertex $v_0$ of the icosahedron (see Fig.~\ref{fig_M2}). Consequently, $\frac{(12-1)!}{5}$ permutations of the set $\{1, \dots, 12\}$ were tested since Constraint~(\ref{constraint7}) fixes the value of $v_0$ and four-fifths of the symmetries are removed. The total CPU time is 7.03e5 s (8.13 days). Thus, all permutations of $\{1, \dots, 12\}$ arranged on the 12 vertices of the icosahedron can lead to at least one ADTS.

%% file: src/Conclusion.tex
\section{Conclusions and Future Work}\label{conclusion}
By means of constraint programming approach, we have proved the existence of the ADTS, and any permutation of $\left\{1 \ldots 12 \right\}$ for the vertices of the icosahedron can produce at least one ADTS. But the rest of questions posed in the Introduction (Sec.~\ref{intro}) remain open to us. Even the third question requires non-trivial efforts.  Because when we enforce a set of 20 types of triangular, which is chosen from the 24 triangular tiles, on the model, the upper bound of traversing the entire search tree is $4^{60}$ if we do not take account of constraint propagation. Hence, to find the 20 different tiles whose corresponding ADTSs cover 
all the permutations of $\left\{1 \ldots 12\right\}$ arranged on the vertices of the icosahedron (Question 4) is even more difficult. 

As future work, we plan to employ parallel constraint solving to seek to answer the rest of the questions. Furthermore, we believe that Question 4 (Sec.~\ref{intro}) requires a well-designed nogood recording mechanism to avoid exploring the search space including the permutations of $\left\{1 \ldots 12\right\}$ already visited.
Finally, we propose that the icosoku problem can be a standard benchmark for CSPLib~\cite{csplib}, which is a library of test problems for constraint solvers. Because we believe it has the advantages as an excellent CSP benchmark should have, which are summarized as follows: (1) The constraints required by the CSP model of the benchmark are widely implemented in the state-of-art constraint solvers. (2) The benchmark can be readily generalized and scaled from easy instance to difficult instances, e.g., increasing the number of golfers from 15 to 18 results in the Social Golfer Problem~\cite{csplib_prob010} more difficult to be solved~\cite{icaart19Ke}. Indeed, the regular icosahedron has the highest number of faces among the five Platonic solids, which limits its scalability to increase the difficulty of the problem. But the problem can be expanded to other polyhedra such as Kepler–Poinsot polyhedron or higher dimensions (e.g., four-dimensional Platonic Solids). (3) The benchmark does not rely on third party data (e.g., the Travelling Salesman Problem needs maps of instances), which is more convenient to make comparisons.
\newline
\newline
\textbf{Acknowledgment.}~We should like to thank our colleague Ekkehard K{\"o}hler for drawing our attention to the icosoku problem.